\documentclass{llncs}
\usepackage{amsmath,graphicx,amssymb}
\usepackage{color,subfigure,url}
\usepackage{flushend}
\input{psfig.sty}

\usepackage{fancyheadings}
\begin{document}

\mainmatter

\title{Stacked Autoencoders for  Medical Image Search}
\author{S. Sharma\inst{1}, I. Umar\inst{1},  L. Ospina\inst{1}, D. Wong\inst{1}, H.R. Tizhoosh\inst{2}}
\institute{Systems Design Engineering,University of Waterloo,Canada \\
\and KIMIA Lab,University of Waterloo,Canada \email{tizhoosh@uwaterloo.ca}}


\maketitle
\begin{abstract}
Medical images can be a valuable resource for reliable information to support medical diagnosis. However, the large volume of medical images makes it challenging to retrieve relevant information given a particular scenario. To solve this challenge, content-based image retrieval (CBIR) attempts to characterize images (or image regions) with invariant content information in order to facilitate image search. This work presents a feature extraction technique for medical images using stacked autoencoders, which encode images to binary vectors. The technique is applied to the IRMA dataset, a collection of 14,410 x-ray images in order to demonstrate the ability of autoencoders to retrieve similar x-rays given test queries. Using IRMA dataset as a benchmark, it was found that stacked autoencoders gave excellent results with a retrieval error of 376 for 1,733 test images with a compression of 74.61\%.

\end{abstract}

\section{Introduction}
Many physicians have experienced lawsuits due to perceived or actual medical malpractice and negligence in recent decades. Radiology, due to its diagnostic nature, has been one of most liable branches in medicine. Most claims and complaints deal with a correct diagnosis  \cite{kinsey2015interfacing}. For instance, wrong interpretation of a malignant mass as a benign lesion accounts for approximately 45\% of radiologists' errors \cite{hl1978visual}. In addition, medical claims of misdiagnosis from medical images in Canada cost taxpayers over $8.3$ million per year \cite{association20162014}.

Radiologists do not have a reliable tool to cross-reference their initial diagnosis instantaneously. Receiving a second opinion on the diagnosis requires consulting a peer who must be pyhsically or virtually available, which is not always possible and constitutes an insurmountable financial and personal challenge for hospitals and clinics around the globe. Building a search engine for fast and accurate content-based image retrieval (CBIR) can potentially revolutionize diagnostic radiology and ultimately decrease both the number of patients affected by misdiagnosis and the number of costly claims against clinicians and hospitals. There are untapped petabytes of data already available in digital archives of hospitals -- constantly growing --, that can be leveraged to reduce misdiagnosis through retrieving similar cases to exploit available knowledge from the past.

The challenge CBIR is that the image must be examined at the pixel level to quantify the similarity between images which is not an inherently quantitative attribute, especially in medical imaging. Additionally, when dealing with ``\emph{big image data}'', the computational costs of retrieval can become infeasible, if the image representation, i.e., image feature, is not chosen appropriately. Accuracy and speed of retrieval are understandably crucial in medical image applications. This paper looks at medical image search based on global image similarity. Single layered and stacked autoencoders are examined to search for similar images within the IRMA dataset that contains 14,410 x-ray images.  Similarity is measured using the IRMA error score. 
The idea proposed is  that given an input x-ray image, the system would search for the top $n=3,5,\dots$  most similar images to display to the clinicians (ideally accompanied with other corresponding information such as biopsy reports, treatment plans, monitoring and follow-ups). This would allow practitioners to leverage the diagnosis, monitoring and treatment results of other patients with similar cases.

\section{Background}
\label{sec:BKG}

The main idea proposed in this paper is feature extraction through the use of autoencoders. The main purpose of extracting features from x-ray images is to create a high-level description from low-level pixel values data.  The accuracy of image retrieval is highly dependent on the quality of the features extracted. If the features do not represent the image content adequately, similar images cannot be retrieved.

Image search and retrieval has been studied over the past 20 years \cite{C3}. CBIR focuses on visual information as opposed to textual metadata, which is text-based search as we all know from daily Internet search when we type keywords to search for desired webpages  \cite{C5}. However, implementing autoencoders as a feature extraction technique for x-rays has only been studied recently.  CBIR is a valuable option for medical images. Text-based search may be limited for medical images due to insufficient text-based data or features, as well as lack of standardized software and procedure in clinical enviroments. Akg{\"u}l et al. examine the various applications of CBIR towards medical imaging \cite{C1}. 

Feature extraction is the process of consolidating high signal information from the noisy input, in this case an image. These features are valuable in a model for training and retrieval. Some examples of visual features are color, shape, and texture, which can be extracted from low-level pixel information  \cite{F1}. These visual features can range from low-level descriptors that are focused on a pixel or small group of pixels, to mid-level features that describe shapes, textures and colour, and finally to high-level features that describe the image and its components in rather general or abstract ways. 

Tradition feature extractions have been extensively examined in computer vision. Using artificial neural networks to ``represent'' an image is rather a new development in computer vision. An autoencoder is a special type of neural network that automatically finds compressed encodings of images, while minimizing error. The compressed image can then be used to represent the image to search algorithms. A stacked autoencoder inputs the results from the previous autoencoder. Hinton et al. introduced the idea of using backpropagation networks to develop a complex unsupervised learning scheme \cite{L1}. The autoencoders were trained through Deep Belief Networks \cite{L1}. A denoising autoencoder can be trained on corrupted input versions in order to reduce the impact of future input noise \cite{L2}. Autoencoders have also been applied to mammograms for compression \cite{L3}. Ideas have been proposed to detect (and to ignore) irrelevant image blocks in each medical image class, by analyzing the error histogram of a autoencoder \cite{camlica2015autoencoding}. The purpose in such approaches is to reduce the dimensionality of features for image retrieval when dealing with a large number of images. The relevance of image blocks is directly proportional to the error of an autoencoder \cite{camlica2015autoencoding}. However, this approach only uses shallow autoencoders that mostly depend on features used to create the prediction model. On the other hand, the use of deep autoencoders allow to extract a better representation of the raw features of the image, therefore allowing to create a more accurate retrieval scheme. Some considerations in evaluating the suitability of autoencoders for this domain were the accuracy of retrieving similar images. Providing fast and accurate results to practitioners is necessary. Autoencoders may require longer training time dependent on the dimensionality of the problem, however this does not impact the retrieval time, once all images have been processed. 

\section{Proposed Approach}
Fast and accurate CBIR is possible by extracting features from images that have the following two intuitive properties: two very similar images should produce two very similar feature vectors, and two very dissimilar images should produce two very different feature vectors. These two properties are especially critical for medical imaging where a region of interest needs to be analyzed (e.g., a tumour, an organ, a tissue type). A system is only valuable if it can return cases which contain regions similar to the selected region of interest.

Some CBIR solutions use local feature from feature descriptors such as SURF \cite{velmurugan2011contentbased}. This technique creates many local feature descriptors for a given image and then combines all of them (up to a threshold) to create a pseudo-global image descriptor. This approach works well for general images but, as we verified in many preliminary experiments, fails in medical imaging, specifically for x-ray images. This could be due to x-rays not having many sharp or ``good'' features, e.g., corners, for the algorithms to use. To circumvent this problem, a global first approach is to be used to create one global feature descriptor (vector) for a given image. One such approach is to use autoencoders to generate a compressed feature vector. 

Autoencoders, with $n/p/n$ architecture, encodes $n$ inputs into $p$ positions, and then decode $p$ positions back into $n$ outputs. By setting $p<n$, the autoencoder functions as a compressor to reduce the dimensionality. Such an autoencoder is basically a shallow neural network with some level of error to reconstruct the input signal from its minimalistic representation in the deepest layer.

While it is possible to use many visual features such as color, shape and texture as part of a feature extraction pipeline, we propose a very simple scheme. We use the unprocessed pixel values as features into the autoencoder. The autoencoder, by virtue of its dimensionality reduction capability, will extract the complex anatomical features using the reconstruction error as its guide. We hypothesize that these features will satisfy the two properties mentioned above since the features are constructed relative to the training set, a claim that we seek to experimentally establish in following sections. The architecture can be deepened by creating an architecture of form  $n/m/k/p/k/m/n$ (5 hidden layers) with $n\!<\!m\!<\!k\!<\!p$ by stacking multiple single layer autoencoders serially. Stacked autoencoders (SAE) can capture highly nonlinear mapping between input and output from the interactions between the many hidden layers and thousands of trainable weights.

\section{Experiments}
In this section, we describe the IRMA dataset, report the details of training and testing the stacked autoencoder, and finally analyze the results of image retrieval using autoencoders applied on IRMA dataset.
  
\subsection{Image Data}
Our goal is to provide the $n$ most similar image to an trained medical professional, the clinician, when she/he provides a query image. Images can be verified to be similar using many techniques such as RMS, and SSIM, however these metrics do not reflect medical domain information. To verify if two x-rays are indeed similar, the images would need to be shown to a medical professional. The IRMA data set contains images already annotated by medical professionals, which makes it a very useful benchmark dataset for retrieval purposes.
 
\begin{figure}[htb]
\centering     
\subfigure[\tiny 1121-127-700-500]{\label{fig:a}\includegraphics[width=2.9cm,height=2.9cm]{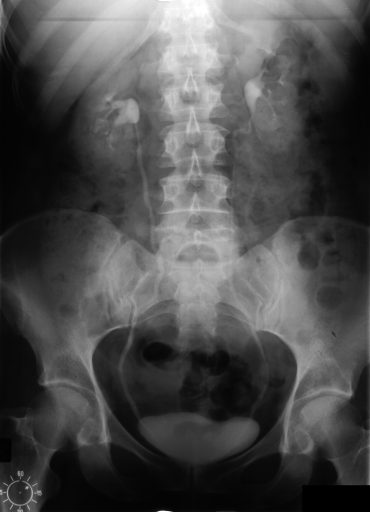}}
\subfigure[\tiny 1121-120-942-700]{\label{fig:b}\includegraphics[width=2.9cm,height=2.9cm]{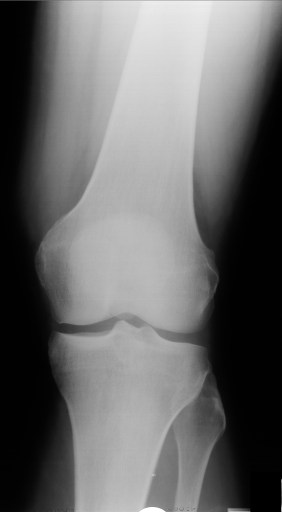}}
\subfigure[\tiny 1121-120-918-700]{\label{fig:b}\includegraphics[width=2.9cm,height=2.9cm]{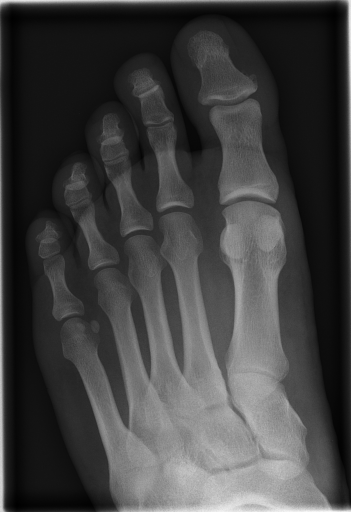}}
\subfigure[\tiny 1121-240-442-700]{\label{fig:b}\includegraphics[width=2.9cm,height=2.9cm]{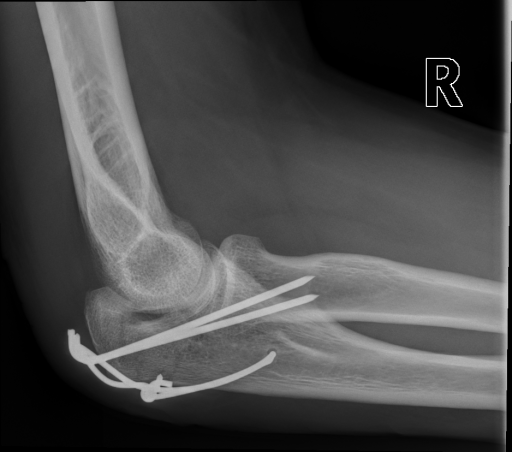}}\\
\subfigure[\tiny 1121-120-200-700]{\label{fig:b}\includegraphics[width=2.9cm,height=2.9cm]{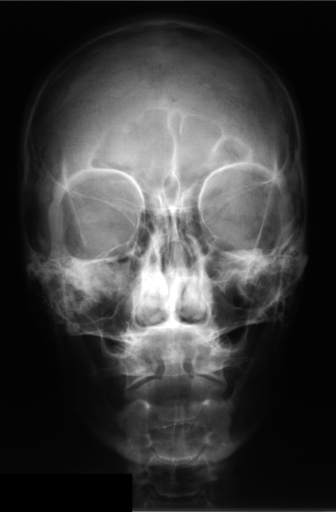}}
\subfigure[\tiny 1123-127-500-000]{\label{fig:b}\includegraphics[width=2.9cm,height=2.9cm]{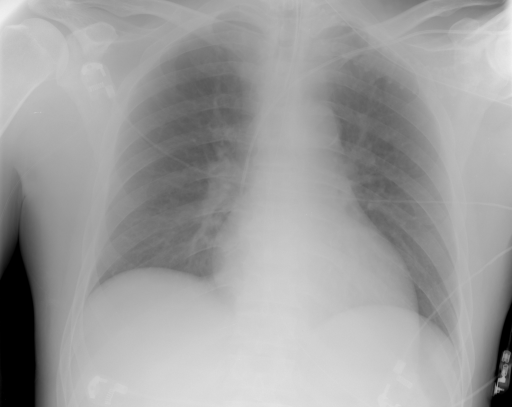}}
\subfigure[\tiny 1121-220-310-700]{\label{fig:b}\includegraphics[width=2.9cm,height=2.9cm]{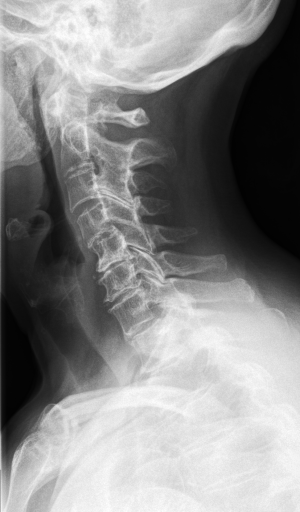}}
\subfigure[\tiny 112d-121-500-000]{\label{fig:b}\includegraphics[width=2.9cm,height=2.9cm]{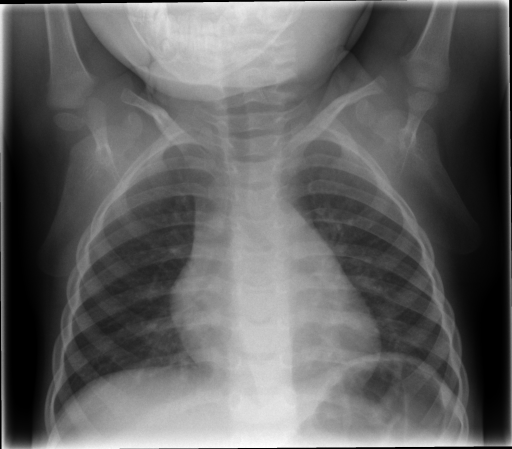}}
\caption{Sample images with their IRMA codes TTTT-DDD-AAA-BBB.}
\label{fig:IRMASamples}
\end{figure}

The IRMA dataset, supplied for the imageCLEF organization, had annotated codes were similar to domain expert knowledge \cite{3,Lehmann2003,Lehmann2006,Mueller2010}. This database has been used by many researchers and is comprised of preset  test and training portions, which enable direct comparisons of metrics. The IRMA codes (manually created by several clinicians) contains information on technical, biological and diagnostic traits of the image in a structured manner: TTTT-DDD-AAA-BBB. Each section is hierarchical meaning there is a least significant bit and most significant digit (Table \ref{tab:irma}). Sample IRMA images are depicted in Fig. \ref{fig:IRMASamples}. 

\begin{table}[htp]
\caption{Sections of the IRMA code}\vspace{0.03in}
\begin{center}
\begin{tabular}{|c|l|}
\hline
T & Image modality and direction \\ \hline
D & Body orientation and anatomical code \\ \hline
A & Region of body examined \\ \hline
B & Biological system examined \\ \hline
\end{tabular}
\end{center}
\label{tab:irma}
\end{table}%

As mentioned before, the objective of any CBIR system is to retrieve the most similar image from a dataset given an input (query) image. To build the database, each image in the training dataset is encoded using a trained Autoencoder. Processing the image through the autoencoder results in a compressed (encoded) feature vector that is assumed to ``describe'' the image content. This feature vector is created for all 12,998 x-ray images of IRMA dataset and stored. To retrieve an image, the query image is put through the trained autoencoder and a query feature vector is generated. The feature vector is then compared with all other indexed feature vectors to locate the most similar image. Various autoencoder specifications were tried, their results are stated in following sections.

\textbf{IRMA Error Score --} Let $I=\hat{l}_1, \hat{l}_2, \dots, \hat{l}_i, \dots, \hat{l}_I$ be the classified IRMA code (for one axis: T,D,A or B) of an image; where $l_i$ is set precisely for every position, and in $\hat{l}i$ is used when we `don't know', marked  by `*'. $I$ may be different for different images. If the classification at position $\hat{l}_i$ is wrong, then all succeeding decisions are wrong and, given a not-specified position, all succeeding decisions are considered to be not specified. if the correct code is unspecified and the predicted code is a wildcard, then no error is counted. In such cases, all remaining positions are regarded as not specified. Wrong (easy) decisions (i.e., fewer possible choices) are penalized for wrong difficult decisions (many possible choices at that node). A decision at position $l_i$ is correct by chance with a probability of $\frac{1}{b_i}$ if $b_i$ is the number of possible labels for position $i$. This assumes equal priors for each class at each position. Furthermore, wrong decisions at an early stage in the code (higher up in the hierarchy) are more penalized than wrong decisions at a later stage in the code (lower down on the hierarchy): i.e., $l_i$ is more important than $l_{i+1}$. Considering all these thoughts: 
\begin{equation}
\textrm{IRMA Error} = \sum_{i=1}^{I} \frac{1}{b_i} \frac{1}{i} \delta(I_i,\hat{I}_i) 
\end{equation}
with $\delta(I_i,\hat{I}_i)$ being 0, 0.5 or 1 for agreement, `don't know' and disagreement, respectively. In order to normalize the error, the maximal possible error is calculated for every axis  such that a completely wrong decision (i.e., all positions for that axis are wrong) gets an error count of 0.25 and a completely correctly predicted axis has an error of 0. Therefore, an image with all positions in all axes being wrong has an error count of 1, and an image with all positions in all axes being correct has an error count of 0. Below a sample calculation of an error between a query and matched image:

\vspace{0.07in}
IRMA code of Query Image: $~~~~~~1111-223-555-777$

IRMA code of Retrieved Image: $~1111-010-555-778$ 

Digits Wrong: $\qquad\qquad\qquad~~~~~~~0000-111-000-001$

Score (normalized): $\qquad\qquad\quad~~~0.2835$

\vspace{0.07in}

The result is normalized on a scale 0 to 1. The IRMA error score is accumulated over the entire test set (1,734 images). The resulting number is the IRMa score reported in the results section. Since each IRMA error can range from 0 to 1, the sum of all error divided by the total possible error gives the error percentage for the test set. 

\subsection{Architecture of Autoencoder}
We investigate both single layer autoencoders $n/p/n$ where $p \ll n$ and stacked autoencoders where the output of the previous autoencoder is fed to the next and so on. For the actual search we use the $k$-NN algorithm. 

\textbf{Training single layer autoencoder -- } We take 12,998 training x-ray images from the IRMA dataset. Each image is gray-scaled, and downscaled to $32 \times 32$ and normalized in $[0, 1]$. It is then fed to the input layer of the autoencoder as a column vector of size 1024. The encoding and decoding layers are trained to reconstruct the input vector (down sampled and vectorized image) by minimizing the reconstruction error measured by cross entropy using stochastic gradient descent. Note that we are using tied weights for encoding and decoding weights which has a regularization effect. We use sigmoid function for nonlinearity. Finally, we obtain the feature vector for each image by obtaining the latent representation from the hidden layer for each image.

\begin{figure*}[tb]
\begin{center}
\includegraphics[width=\textwidth]{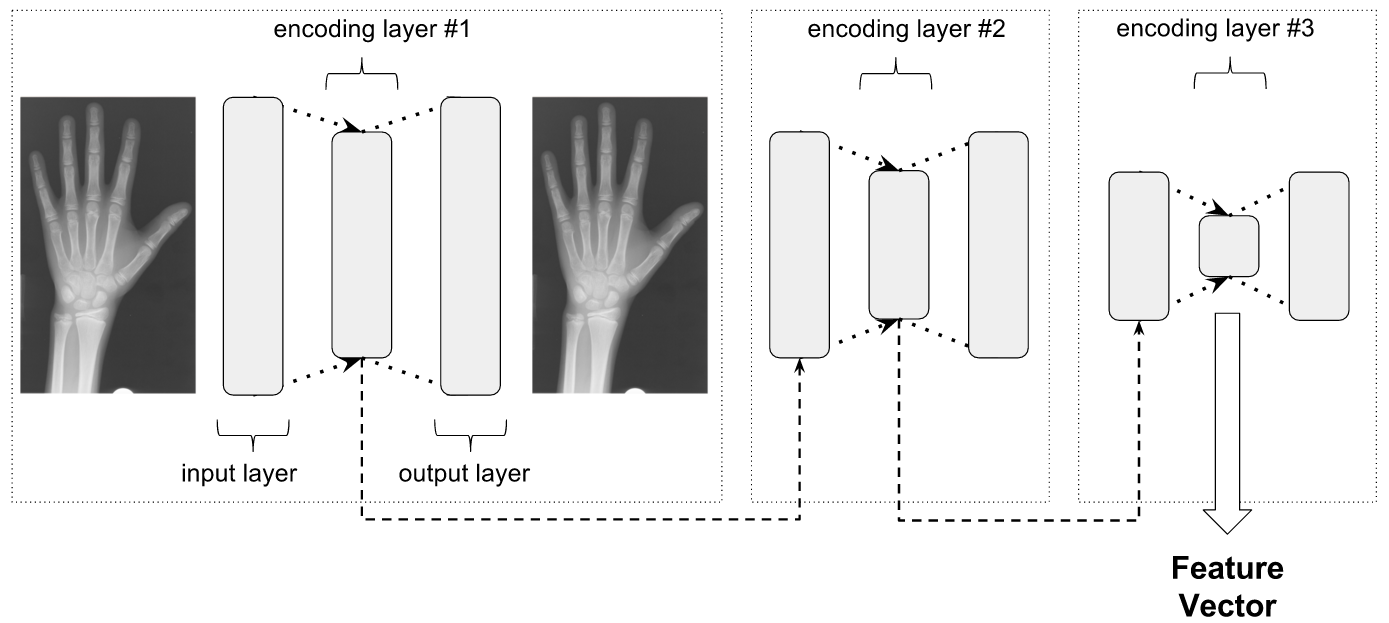}
\caption{Three hidden layers constructed via stacked autoencoders where the feature vector is extracted from the deepest layer.}
\label{fig:auto}
\end{center}
\end{figure*}

\textbf{Training stacked autoencoder --} We obtain stacked autoencoders by serially arranging autoencoders such that the output of the first layer is fed as input to the second and so on (Fig. \ref{fig:auto}). Each autoencoder layer is trained greedily. Latent features from the first autoencoder are used as input to the second autoencoder layer. The process is repeated for all subsequent hidden layers. Finally, the latent feature vector from the last encoding layer acts as the final feature vector for the input image. We use Theano \cite{T1,T2} and various python libraries to implement both the autoencoder and stacked autoencoder \cite{S1}.

\subsection{Results}
We report compression reduction, IRMA score and reconstruction error (root mean squared) for various AEs (autoencoders) and SAEs (stacked autoencoders).

We tried various configurations for a single layer autoencoder where $p < n$ as reported in Table \ref{tab:single}. We used Euclidean distance to compare feature vectors between two images. The first hit (1-NN) IRMA score as well as the reconstruction error is reported on the test set. Interestingly, better reconstruction error did not necessarily lead to lower IRMA score. For example, while the 1024/512/1024 configuration clearly had the lowest reconstruction error amongst the reported single layer architectures, it did not yield the lowest IRMA score. This implies that the features trained by minimizing the reconstruction error heuristic are clearly useful, we cannot easily predict the performance of the features on the test dataset from the reconstruction error alone. We note that the latent features produced significant compression while still maintaining low error. Each trial used the following settings: 30 epochs, a learning rate of 0.1, and a batch size of 20 for stochastic gradient descent.

\begin{table}[ht]
\caption{Results for an autoencoder with a single hidden layer (RMS captures the test reconstruction error). }
\begin{center}
\begin{tabular}{|l|c|c|c|}
Architecture & \% Reduction & IRMA score & RMS \\ \hline
1024/225/1024 & 78.03\% & 407 & 0.104277 \\
1024/512/1024 & 50\% & 405 & 0.091310 \\
1024/150/1024 & 85.35\% & 393 & 0.114780 \\
1024/275/1024 & 73.14\% & 388 & 0.100892 \\ \hline
\end{tabular}
\end{center}
\label{tab:single}
\end{table}%

We also include results from stacking two and three autoencoders sequentially (see Tables \ref{tab:two} and \ref{tab:three}). The reconstruction errors are reported on the deepest layer. The 1024/600/1024, 600/500/600, 500/260/500 architecture yielded the lowest IRMA score of 376 while still maintaining relatively high compression (74.61\% reduction). We believe stacked autoencoders, as opposed to a single layer autoencoder, produce a more useful higher-level representation from the lower-level representation output by the previous layer.  To test for overfitting on the training data, the reconstruction error was calculated on the training set and compared to the testing images. For example, a triple stacked autoencoder 1024/600/1024, 600/500/600, 500/260/500, the error was 0.096200 and 0.101679, respectively. One can interpret the negligible difference in reconstruction errors as acceptable generalization. It should be noted that reconstruction error for stacked autoencoders is reported only for the first layer, and therefore the true reconstruction error is not simple to obtain. However, the differences in reconstruction for a simple non-stacked autoencoder (see Table \ref{tab:three}) are also negligible. 

\begin{figure}[tbh!]
\begin{center}
\includegraphics[width=0.49\columnwidth]{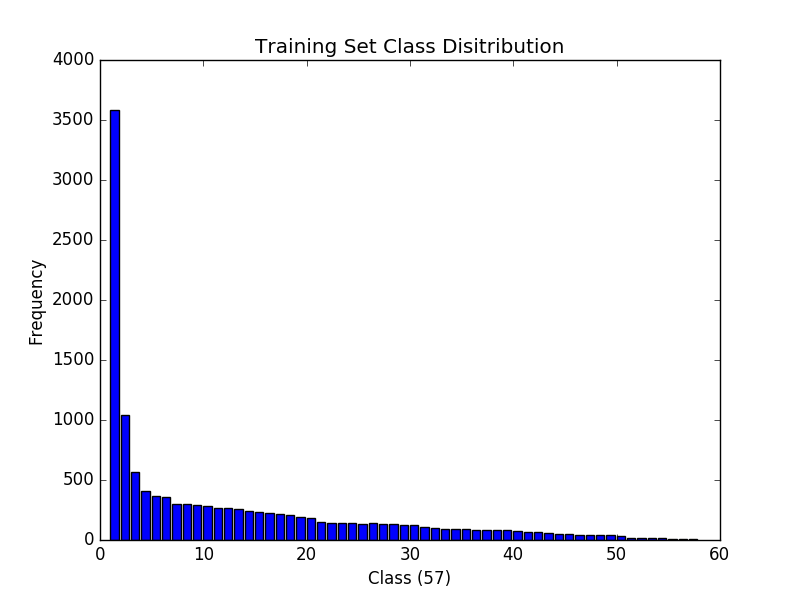}
\includegraphics[width=0.49\columnwidth]{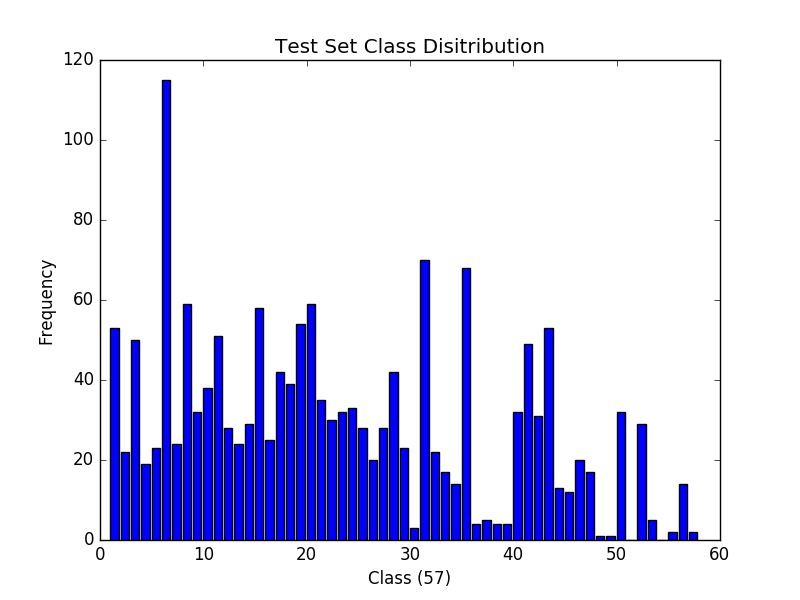}
\caption{Imbalance: class distribution of IRMA training (left) and test data (right). }
\label{fig:dist}
\end{center}
\end{figure}

Observing the distribution of data in the IRMA training and testing images, we noticed that there exists both inter- and intra-statistical imbalances between the data sets. That is, the training set is severely imbalanced such that various classes are not represented normally. A histogram of image classes is shown in Fig. \ref{fig:dist}.  Despite this problem, the trained autoencoders have a low IRMA score and exhibit negligible differences in reconstruction error (between training and test data). This can further compound the notion that there is no overfitting, hence the autoencoders are generalizing well. 

\begin{table*}[ht]
\caption{Results for two hidden layers (TRE=Test Reconstruction Error).}
\begin{center}
\begin{tabular}{|l|c|c|c|}
Layer architecture & \% Reduction & IRMA score & TRE (RMS) \\ \hline
1024/600/1024, 600/250/600 & 75.59\% & 393 & 0.102615 \\
1024/400/1024, 400/250/400 & 75.59\% & 391 & 0.102615 \\
1024/512/1024, 512/250/512 & 75.59\% & 389 & 0.102615 \\  \hline
\end{tabular}
\end{center}
\label{tab:two}
\end{table*}%

\begin{table*}[ht]
\caption{Results for three hidden layers (TRE=Test Reconstruction Error).}
\begin{center}
\begin{tabular}{|l|c|c|c|}
Layer architecture & \% Reduction & IRMA score & TRE (RMS) \\ \hline
1024/600/1024, 600/400/600, 400/200/400 & 80.47\% & 400 & 0.111440 \\
1024/600/1024, 600/500/600, 500/260/500 & 74.61\% & 376 & 0.101679 \\  \hline
\end{tabular}
\end{center}
\label{tab:three}
\end{table*}%

\section{Conclusions}
Content-based image retrieval is a challenging problem for medical images. It is possible to analyze one image in a short amount of time and classify it in a medically meaningful way. However, given a new image, comparing it to the entire collection of medical images, to find the similar images, can be computationally expensive. Such a task requires a technique that searches through the big image data efficiently while still finding highly similar images. One method to reduce the dimensionality of the problem is to encode the images in a smaller, compact representation to enable faster image comparisons during the retrieval task. 
Such technology has clearly enormous potential in aiding domains such as diagnostic radiology and pathology. Due to the scarcity of text-based features associated with medical images directly embedded in medical image formats, i.e., in DICOM files, as well as the large quantity of images available, a search focused on visual features was explored in this paper. A global, content-based image retrieval technique was successfully applied to the application of medical images. The best performing feature extraction technique examined was an stacked autoencoder. Of the examined architectures, the most accurate setup consisted of three hidden layers with configurations 1024/600/1024, 600/500/600, and 500/260/500. On the IRMA dataset, with 14,410 x-ray images, the proposed approach achieved an IRMA score of 376 with a compression of 74.61\%. 

\bibliographystyle{plain}
\vspace{0.08in}
\bibliography{refs} 

\end{document}